\documentclass[10pt,twocolumn,letterpaper]{article}

\usepackage{iccv}
\usepackage{times}
\usepackage{epsfig}
\usepackage{graphicx}

\usepackage{color}
\usepackage{xcolor}
\usepackage{comment}

\usepackage{amsmath,amssymb,amsfonts}
\usepackage{algorithmic}
\usepackage{graphics}
\usepackage{textcomp}
\usepackage[export]{adjustbox}
\usepackage{multicol}
\usepackage{tabularx}
\usepackage{color}

\usepackage[breaklinks=true,bookmarks=false]{hyperref}
\hypersetup{
    colorlinks=true,
    linkcolor=blue,
    filecolor=magenta,      
    urlcolor=red,
}
\iccvfinalcopy 

\def\iccvPaperID{****} 

\ificcvfinal\fi

\begin{document}

\title{Analysing Affective Behavior \\ in the second ABAW2 Competition}

\author{Dimitrios Kollias\\
University of Greenwich, UK\\
{\tt\small D.Kollias@greenwich.ac.uk}
\and
Irene Kotsia \\
Middlesex University London, UK 
\and
Elnar Hajiyev \\  
Realeyes 
\and
Stefanos Zafeiriou \\ 
Imperial College London, UK     

}

\maketitle
\ificcvfinal\thispagestyle{empty}\fi

\begin{abstract}
The Affective Behavior Analysis in-the-wild (ABAW2) 2021 Competition is the second -following the first very successful ABAW Competition held in conjunction with IEEE FG 2020- Competition that aims at automatically analyzing affect. ABAW2 is split into three Challenges, each one addressing one of the three main behavior tasks of Valence-Arousal Estimation, seven Basic Expression Classification and twelve Action Unit Detection. All three Challenges are based on  a common benchmark database, Aff-Wild2, which is a large scale in-the-wild database and the first one to be annotated for all these three tasks.
In this paper, we describe this Competition, to be held in conjunction with ICCV 2021. We present the three Challenges, with the utilized Competition corpora. We outline the evaluation metrics and present the baseline system with its results. More information regarding the Competition is provided in the Competition site: \url{https://ibug.doc.ic.ac.uk/resources/iccv-2021-2nd-abaw/}.
\end{abstract}

\section{Introduction}

The proposed Workshop tackles the problem of affective behavior analysis in-the-wild, which is a major targeted characteristic of HCI systems used in real life applications. The current 5th societal revolution aims at merging the physical and cyber spaces, providing services that contribute to people's well-being. The target is to create machines and robots that are capable of understanding people's feelings, emotions and behaviors; thus, being able to interact in a 'human-centered' and engaging manner with them, and effectively serving them as their digital assistants.

Affective behavior analysis in diverse environments, such as in people's homes, in their work, operational or industrial environments, will have a positive societal impact. It will provide machines and robots with the ability to interact and assist people in an effective and natural way. Through human affect recognition, the reactions of the machine, or robot, will be consistent with people's expectations and emotions; their verbal and non-verbal interactions will be positively received by humans. Moreover, this interaction should not be dependent on the respective context, nor the human's age, sex, ethnicity, educational level, profession, or social position. As a result, the development of intelligent systems able to analyze human behaviors in-the-wild can contribute to generation of trust, understanding and closeness between humans and machines in real life environments.

\begin{figure}[h]
\centering
\adjincludegraphics[height=5.3cm]{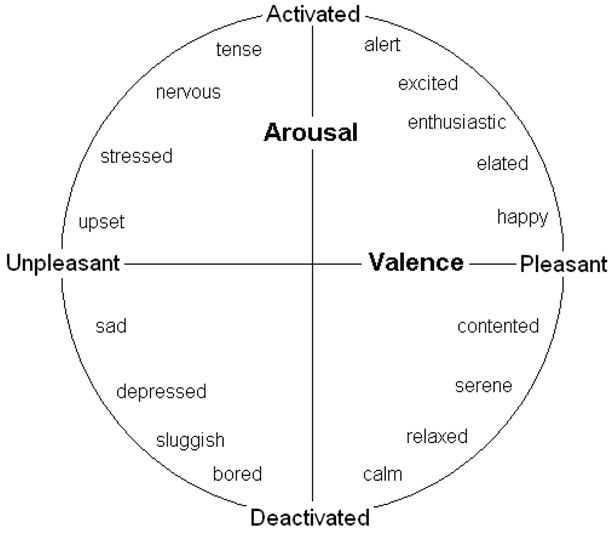}
\caption{The 2D Valence-Arousal Space}
\label{va-space}
\end{figure}

Representing human emotions has been a basic topic of research in psychology. The most frequently used emotion representation is the categorical one, including the seven basic categories, i.e., Anger, Disgust, Fear, Happiness, Sadness, Surprise and Neutral \cite{ekman2003darwin}. Discrete emotion representation can also be described in terms of the Facial Action Coding System (FACS) model, in which all possible facial actions are described in terms of Action Units (AUs) \cite{ekman2002facial}. 
Finally, the dimensional model of affect \cite{whissel1989dictionary,russell1978evidence} has been proposed as a means to distinguish between subtly different displays of affect and encode small changes in the intensity of each emotion on a continuous scale. The 2-D Valence and Arousal (VA) Space (valence shows how positive or negative an emotional state is, whereas arousal shows how passive or active it is) is the most usual dimensional emotion representation, depicted in Figure \ref{va-space}. 

There are a number of related applications spread across a variety of fields, such as medicine, health, driver fatigue, monitoring, e-learning, marketing, entertainment, lie detection, law
\cite{tagaris1,acharya2018automated,kollias13,kim2016deep,mdpi,tagaris2,springer,nasser2019artificial,Tailor,kollias2021mia,caridakis10synthesizing,tzouveli2008adaptive,wallace2005intelligent,malatesta2009towards,rapantzikos2007spatiotemporal,tsapatsoulis2000fuzzy,anastassiou1988digital,kollias1988adaptive,glimm2013using,kollia2011answering,kollia2012improving}.

The ABAW2 Competition contains three Challenges, which are based on the same database; these target (i) dimensional affect recognition (in terms of valence and arousal) \cite{khorrami2016deep,kollias2018old,chen2017multimodal,kollias2018photorealistic,weichi,kollias2020exploiting,zhang2020m,kollias2020va,kolliasijcv,barros2018omg}, (ii) categorical affect classification (in terms of the seven basic expressions) \cite{ng2015deep,ding2017facenet2expnet,kollias2015interweaving,zhao2016peak,kollias2016line,jung2015joint,kollias2017adaptation,liu2020emotion,kollias6} and (iii) 12 facial action unit detection \cite{pahl2020multi,kuhnke2020two,kollias2021distribution,han2016incremental,kollias2019face,deng2020fau}, in-the-wild. These Challenges produce a significant step forward when compared to previous events.
In particular, they use the Aff-Wild2 \cite{kollias2020analysing,kolliasexpression,kollias2021affect,kollias2018aff2,kollias2018multi}, the first comprehensive benchmark for all three affect recognition tasks in-the-wild: the Aff-Wild2 database extends the Aff-Wild \cite{kollias2018deep,zafeiriou,zafeiriou1}, with more videos and annotations for all behavior tasks. Aff-Wild consists of 298 videos, displaying reactions of 200 subjects, with a total video duration of about 30 hours, and 1,250,000 video frames, annotated in terms of valence and arousal. It has been used in the Aff-Wild Challenge in CVPR 2017, with participation of more than 10 research groups. To generate Aff-Wild2, we added 266 more videos, displaying the reactions of 266 more subjects, with a duration of more than 18 hours, and 1,500,000 frames. Aff-Wild2 includes extended spontaneous facial behaviors in arbitrary recording conditions and a significantly increased number of different subjects (466; 280 of which are males and 186 females) and frames (around 2,800,000).

The remainder of this paper is organised as follows. We introduce the Competition corpora in Section \ref{corpora}, the Competition evaluation metrics in Section \ref{metrics}, the developed baseline per Challenge, along with the obtained results in Section \ref{baseline}, before concluding in Section \ref{conclusion}.


\section{Competition Corpora}\label{corpora}


The second Affective Behavior Analysis in-the-wild (ABAW2) Competition relies on the Aff-Wild2 database \cite{kolliasexpression,kollias2018aff2,kollias2018multi}. Aff-Wild2 is the first ever database annotated for all three main behavior tasks: valence-arousal estimation, action unit detection and basic expression classification. These three tasks form the three Challenges of this Competition. 

Aff-Wild2 consists of $548$ videos with $2,813,201$ frames. Sixteen of these videos display two subjects (both have been annotated). All videos have been collected from YouTube. Aff-Wild2 is an extension of Aff-Wild \cite{kollias2018deep,zafeiriou,zafeiriou1}; 260 more YouTube videos, with $1,413,000$ frames, have been added to Aff-Wild. 
Aff-Wild was the first large scale, captured in-the-wild, dimensionally annotated database, containing 298 YouTube videos that display subjects reacting to a variety of stimuli. 
Aff-Wild2 shows both subtle and extreme human behaviours in real-world settings. The total number of subjects in Aff-Wild2 is 458; 279 of them are males and 179 females.

The Aff-Wild2 database, in all Challenges, is split into training, validation and test set. At first the training and validation sets, along with their corresponding annotations, are being made public to the participants, so that they can develop their own methodologies and test them. The training and validation data contain the videos and their corresponding annotation. Furthermore, to facilitate training, especially for people that do not have access to face detectors/tracking algorithms, we provide bounding boxes and landmarks for the face(s) in the videos (we also provide the aligned faces). At a later stage, the test set without annotations will be given to the participants. Again, we will provide bounding boxes and landmarks for the face(s) in the videos (we will also provide the aligned faces).

In the following, we provide a short overview of each Challenge's dataset and refer the reader to the original work for a more complete description. Finally, we describe the pre-processing steps that we carried out for cropping and aligning the images of Aff-Wild2. The cropped and aligned images have been utilized in our baseline experiments.

\subsection{Aff-Wild2: Valence-Arousal Annotation}

\begin{figure}[h]
\centering
\adjincludegraphics[height=5.6cm]{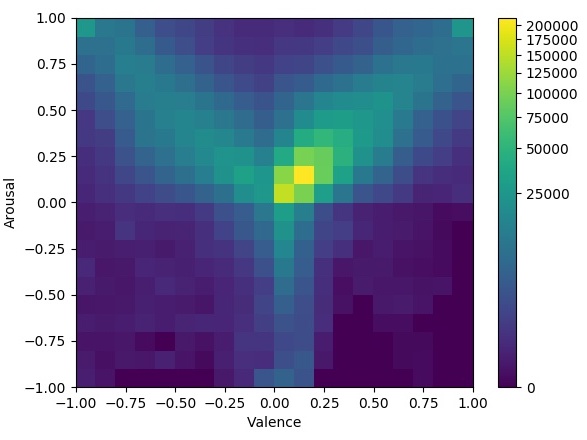}
\caption{2D Valence-Arousal Histogram of Aff-Wild2}
\label{va_annot}
\end{figure}

$545$ videos in Aff-Wild2 contain annotations in terms of valence-arousal. Sixteen of these videos display two subjects, both of which have been annotated. In total,  $2,786,201$ frames, with $455$ subjects, $277$ of which are male and $178$ female, have been annotated by four experts using the method proposed in \cite{cowie2000feeltrace}. The annotators watched each video and provided their (frame-by-frame) annotations through a joystick. A time-continuous annotation was generated for each affect dimension. Valence and arousal values range continuously in $[-1,1]$.  The final label  values  were  the  mean  of  those  four  annotations.   The  mean  inter-annotation correlation is 0.63 for valence and 0.60 for arousal. Let us note here that all subjects present in each video have been annotated. Figure \ref{va_annot} shows the 2D Valence-Arousal histogram of annotations of Aff-Wild2.

Aff-Wild2 is currently the largest (and audiovisual) in-the-wild database annotated for valence and arousal.

Aff-Wild2 is split into three subsets: training, validation and test. Partitioning is done in a subject independent manner, in the sense that a person can  appear  only  in  one  of  those  three  subsets. The  resulting  training, validation and test subsets consist of 346, 68 and 131 videos, respectively; the resulting  training, validation and test subsets contain 5, 3 and 8, respectively, videos that display two subjects. 

\subsection{Aff-Wild2: Seven Basic Expression Annotation}

$539$ videos in Aff-Wild2 contain annotations in terms of the seven basic expressions. Seven of these videos display two subjects, both of which have been annotated. In total, $2,595,572$ frames, with $431$ subjects, $265$ of which are male and $166$ female, have been annotated by seven experts in a frame-by-frame basis. A platform-tool was developed in order to split each video into frames and let the experts annotate each videoframe. Let us mention that in this platform-tool, an expert could score a videoframe as having either one of the seven basic expressions or none (since there are affective states other than the seven basic expressions).

Due to subjectivity of annotators and wide ranging levels of images’ difficulty, there were some disagreements among annotators. We decided to keep only the annotations on which at least six (out of seven) experts agreed. Table \ref{expr_distr} shows the distribution of the seven basic expression annotations of Aff-Wild2.

Aff-Wild2 is currently the largest (and audiovisual) in-the-wild database annotated in terms of the seven basic expressions.

\begin{table}[!h]
\caption{ Number of Annotated Images in Each of the Seven Basic Expressions   }
\label{expr_distr}
\centering
\begin{tabular}{ |c||c| }
\hline
Basic Expression & No of Images \\
\hline
\hline
Neutral & 538,411  \\
 \hline
Anger & 52,005  \\
 \hline
Disgust & 31,138  \\
 \hline
Fear &  26,062 \\
 \hline
Happiness & 395,352  \\
 \hline
Sadness & 173,842  \\
 \hline
Surprise & 99,863  \\
 \hline
\end{tabular}
\end{table} 
 
Aff-Wild2 is split into three subsets: training, validation and test. Partitioning is done in a subject independent manner. The  resulting  training, validation and test subsets consist of 250, 70 and 222 videos, respectively; the resulting  training, validation and test subsets contain 3, 0 and 1, respectively, videos that display two subjects.

\subsection{Aff-Wild2: Twelve Action Unit Annotation}

$534$ videos in Aff-Wild2 contain annotations in terms of twelve action units. Seven of these videos display two subjects, both of which have been annotated. In total, $2,565,169$ frames, with $426$ subjects, $262$ of which are male and $164$ female, have been annotated in a semi-automatic procedure (that involves manual and automatic annotations). Aff-Wild2 has been annotated for the occurrence of twelve action units in a frame-by-frame basis. Table \ref{au_distr} shows the name of the twelve action units that have been annotated, the action that they are associated with and the distribution of their annotations in Aff-Wild2.

Aff-Wild2 is currently the largest (and audiovisual) in-the-wild database annotated in terms of action units.

\begin{table}[h]
    \centering
        \caption{Distribution of AU annotations in Aff-Wild2}
    \label{au_distr}
\begin{tabular}{|c|c|c|}
\hline
  Action Unit \# & Action  &\begin{tabular}{@{}c@{}} Total Number \\  of Activated AUs \end{tabular} \\   \hline    
    \hline    
   AU 1 & inner brow raiser & 294,591 \\   \hline    
   AU 2 & outer brow raiser & 136,569 \\   \hline    
   AU 4 & brow lowerer & 384,969 \\  \hline    
   AU 6 & cheek raiser & 618,929 \\  \hline    
   AU 7 & lid tightener & 618,929 \\  \hline    
   AU 10 & upper lip raiser & 845,793 \\  \hline    
   AU 12 & lip corner puller & 598,699 \\  \hline    
   AU 15 & lip corner depressor & 62,954 \\  \hline    
  AU 23 & lip tightener & 77,793 \\  \hline    
   AU 24 & lip pressor & 61,460 \\  \hline    
   AU 25 & lips part & 1,579,262 \\  \hline     
   AU 26 & jaw drop & 202,447 \\  \hline     

\end{tabular}
\end{table}

Aff-Wild2 is split into three subsets: training, validation and test. Partitioning is done in a subject independent manner. The  resulting  training, validation and test subsets consist of 302, 105 and 127 videos, respectively; the resulting  training, validation and test subsets contain 3, 0 and 4, respectively, videos that display two subjects.

\subsection{Aff-Wild2 Pre-Processing: Cropped \& Cropped-Aligned Images} \label{pre-process}


At first, we split all videos into images (frames). Then, the SSH face detector \cite{najibi2017ssh,tsapatsoulis2000face} based on the ResNet \cite{he2016deep} and trained on the WiderFace dataset \cite{yang2016wider} was used to extract face bounding boxes from all the images. The cropped images according to these bounding boxes were provided to the participating teams.
Also, 5 facial landmarks (two eyes, nose and two mouth corners) were extracted and used to perform similarity transformation. The resulting cropped and aligned images were additionally provided to the participating teams. Finally, the cropped and aligned images were utilized in our baseline experiments, described in Section \ref{baseline}.

\section{Evaluation Metrics Per Challenge}\label{metrics}

Next, we present the metrics that will be used for assessing the performance of the developed methodologies of the participating teams in each Challenge.

\subsection{Valence-Arousal Estimation Challenge} 

\noindent The Concordance Correlation Coefficient (CCC) is widely used in measuring the performance of dimensional emotion recognition methods, such as in the series of AVEC challenges \cite{ringeval2019avec}. CCC evaluates the agreement between two time series (e.g., all video annotations and predictions) by scaling their correlation coefficient with their mean square difference. In this way, predictions that are well correlated with the annotations but shifted in value are penalized in proportion to the deviation. CCC takes values in the range $[-1,1]$, where $+1$ indicates perfect concordance and $-1$ denotes perfect discordance. The highest the value of the CCC the better the fit between annotations and predictions, and therefore high values are desired.
CCC is defined as follows:

\begin{equation} \label{ccc}
\rho_c = \frac{2 s_{xy}}{s_x^2 + s_y^2 + (\bar{x} - \bar{y})^2},
\end{equation}

\noindent
where $s_x$ and $s_y$ are the variances of all video valence/arousal annotations and predicted values, respectively, $\bar{x}$ and $\bar{y}$ are their corresponding mean values and $s_{xy}$ is the corresponding covariance value.

The mean value of CCC for valence and arousal estimation will be adopted as the main evaluation criterion. 

\begin{equation} \label{va}
\mathcal{E}_{total} = \frac{\rho_a + \rho_v}{2},
\end{equation}

\subsection{Seven Basic Expression Classification Challenge}\label{evaluation}

\noindent The $F_1$ score is a weighted average of the recall (i.e., the ability of the classifier to find all the positive samples) and precision (i.e., the ability of the classifier not to label as positive a sample that is negative). The $F_1$ score reaches its best value at 1 and its worst score at 0. The $F_1$ score is defined as:

\begin{equation} \label{f1}
F_1 = \frac{2 \times precision \times recall}{precision + recall}
\end{equation}

The $F_1$ score for emotions is computed based on a per-frame prediction (an emotion category is specified in each frame).

Total accuracy (denoted as $\mathcal{T}Acc$) is defined on all test samples and is the  fraction of predictions that the model got right. Total accuracy reaches its best value at 1 and its worst score at 0. It is defined as:

\begin{equation} \label{total accuracy}
\mathcal{T}Acc = \frac{\text{Number of Correct Predictions}}{\text{Total Number of Predictions}}
\end{equation}

A weighted average between the $F_1$ score and the total accuracy, $\mathcal{T}Acc$, will be the main evaluation criterion:

\begin{equation} \label{expr}
\mathcal{E}_{total} = 0.67 \times F_1 + 0.33 * \mathcal{T}Acc,
\end{equation}

\subsection{Twelve Action Unit Detection Challenge}\label{evaluation2}

To obtain the overall score for the AU detection Challenge,
we first obtain the $F_1$ score for each AU independently, and then compute the (unweighted) average over all 12 AUs (denoted as $\mathcal{A}F_1$) :
\begin{equation} \label{au1}
\mathcal{A}F_1 = \sum_{i=1}^{12} F_1^i
\end{equation}

The $F_1$ score for AUs is computed based on a per-frame detection (whether each AU is present or absent). 

The average between the $\mathcal{A}F_1$ score and the total accuracy, $\mathcal{T}Acc$, will be the main evaluation criterion:

\begin{equation} \label{aus}
\mathcal{E}_{total} = 0.5 \times \mathcal{A}F_1 + 0.5 * \mathcal{T}Acc
\end{equation}

\section{Baseline \& Participating Teams' Systems and Results} \label{baseline}

All baseline systems rely exclusively on existing open-source machine learning toolkits to ensure the reproducibility of the results.
In this Section, we first describe the baseline systems developed for each Challenge and then report their obtained results.

\begin{figure*}[h]
\centering
\adjincludegraphics[width=1\linewidth]{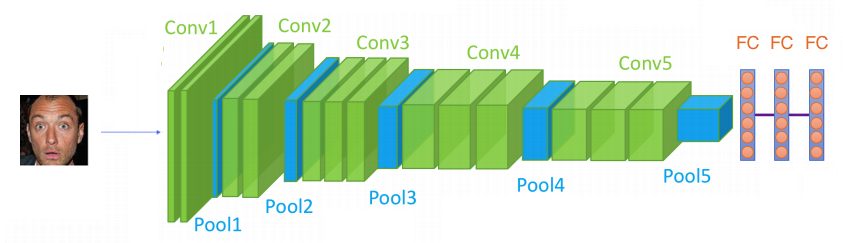}
\caption{The architecture of the utilized baseline VGG-FACE that has been used in all Challenges; the output is either linear (VA case), or with a softmax unit (in the 7 basic expression case), or with a sigmoid unit (in the 12 AU case)}
\label{vgg}
\end{figure*}

At first, let us mention that we utlized the cropped and aligned images from Aff-Wild2, as described in Section \ref{pre-process}. These images have dimensions $112 \times 112 \times 3$. The pixel intensities are normalized to take values in [-1,1]. No on-the-fly or off-the-fly data augmentation technique \cite{kuchnik2018efficient,kollias2018photorealistic,kolliasijcv} was utilized.


\subsection{Baseline Systems}

The architecture that was used in all 3 Challenges was based on the 13 convolutional and pooling layers of VGG-FACE \cite{parkhi2015deep} (its fully connected layers are discarded), followed by 2 fully connected layers, each with
4096 hidden units. In the Valence-Arousal Estimation Challenge baseline, a (linear) output layer follows that gives final estimates for valence and arousal. In the Seven Basic Expression Classification Challenge, a final output layer with softmax as activation function follows which gives the 7 basic expression predictions. In the twelve Action Unit Detection Challenge, a final output layer with sigmoid as activation function follows which gives the 12 action unit predictions. Figure \ref{vgg} shows this basic architecture of the utilized VGG-FACE.

The baseline systems have been pre-trained on the VGG-Face dataset; their convolutional layers were fixed (i.e., non-trainable) and only the three fully connected were trained on Aff-Wild2. These systems have been implemented in TensorFlow; training time was around a day on a Titan X GPU, with a learning rate of $10^{-4}$ and with a batch size of 256.

\subsection{Results}

Table \ref{ccc_results} presents the CCC evaluation of valence and arousal predictions on the Aff-Wild2 validation set, of the baseline network (VGG-FACE). 

\begin{table}[!h]
\caption{Baseline results for VA estimation on the validation set of Aff-Wild2; $\mathcal{E}_{total}$ is the mean valence and arousal CCC}
\label{ccc_results}
\centering
\begin{tabular}{ |c||c|c|c| }
\hline
\multicolumn{1}{|c||}{Baseline}  & \multicolumn{2}{c|}{CCC} & \multicolumn{1}{c|}{$\mathcal{E}_{total}$} \\
\hline
      & Valence & Arousal &   \\
 \hline
\hline
VGG-FACE  & 0.23 & 0.21  & 0.22   \\
 \hline
\end{tabular}
\end{table}

Table \ref{expr_results} presents the performance on the validation set of Aff-Wild2, of the baseline network (VGG-FACE) of the seven basic expression classification Challenge. The performance metric is a weighted average between the F1 score and the total accuracy, as discussed in Section \ref{evaluation}.

\begin{table}[!h]
\caption{Baseline results for the seven basic expression classification on the validation set of Aff-Wild2; $\mathcal{E}_{total} = 0.67 \times F_1 + 0.33 * \mathcal{T}Acc$}
\label{expr_results}
\centering
\scalebox{.95}{
\begin{tabular}{ |c||c|c|c| }
\hline
\multicolumn{1}{|c||}{Baseline}  & \multicolumn{1}{c|}{\begin{tabular}{@{}c@{}}F1 \\ Score \end{tabular}} & \multicolumn{1}{c|}{\begin{tabular}{@{}c@{}}Total \\  Accuracy \end{tabular}} & \multicolumn{1}{c|}{$\mathcal{E}_{total}$} \\
 \hline
\hline
VGG-FACE & 0.30 & 0.50 & 0.366  \\  
 \hline
\end{tabular}
}
\end{table}

Table \ref{aus_results} presents the performance on the validation set of Aff-Wild2, of the baseline network VGG-FACE of the twelve Action Unit Detection Challenge. The performance metric is the average between the F1 score and the total accuracy, as discussed in Section \ref{evaluation2}.

\begin{table}[!h]
\caption{Baseline results for 12 action unit detection on the validation set of Aff-Wild2; $\mathcal{E}_{total} = 0.5 \times \mathcal{A}F_1 + 0.5 * \mathcal{T}Acc$}
\label{aus_results}
\centering
\begin{tabular}{ |c||c|c|c| }
\hline
\multicolumn{1}{|c||}{Baseline}  & \multicolumn{1}{c|}{\begin{tabular}{@{}c@{}}Average \\ F1 Score \end{tabular}} & \multicolumn{1}{c|}{\begin{tabular}{@{}c@{}}Total \\  Accuracy \end{tabular}} & \multicolumn{1}{c|}{$\mathcal{E}_{total}$} \\
 \hline \hline
VGG-FACE & 0.40 & 0.22  & 0.31  \\
 \hline
\end{tabular}
\end{table}

\section{Conclusion}\label{conclusion}
In this paper we have presented the second Affective Behavior Analysis in-the-wild Competition (ABAW2) 2020.  It comprises  three Challenges targeting: i) valence-arousal estimation, ii) seven basic expression classification and iii) eight action unit detection. The database utilized for this Competition has been derived from the Aff-Wild2, the large-scale and first database annotated for all these three behavior tasks.
We have also presented the baseline networks and their results.

{\small
\bibliographystyle{ieee_fullname}
\bibliography{egbib}
}

\end{document}